\documentclass[conference]{IEEEtran}
\IEEEoverridecommandlockouts

\usepackage{cite}
\usepackage{amsmath,amssymb,amsfonts}
\usepackage{algorithmic}
\usepackage{graphicx}
\usepackage{textcomp}
\usepackage{xcolor}
\usepackage{comment}
\usepackage{xspace}
\usepackage{hyperref}

\usepackage{listings}
\usepackage[T1]{fontenc}

\usepackage{xcolor}
\usepackage{colortbl}
\usepackage{subcaption}

\usepackage{adjustbox, booktabs, longtable}

\newcommand*{\msteb}{mSTEB}
\newcommand*{\gpt}{GPT-4o}
\newcommand*{\gpta}{GPT-4o Audio}
\newcommand*{\gemini}{Gemini 2.0 Flash}
\newcommand*{\gemma}{Gemma 3 27B}
\newcommand*{\qwen}{Qwen 2 Audio}
\newcommand*{\seamless}{SeamlessM4T v2}
\newcommand*{\lid}{\textsc{LID}}
\newcommand*{\tc}{\textsc{TC}}
\newcommand*{\rc}{\textsc{RC-QA}}
\newcommand*{\nli}{\textsc{NLI}}

\newcommand*{\asr}{\textsc{ASR}}
\newcommand*{\st}{\textsc{S2TT}}

\def\BibTeX{{\rm B\kern-.05em{\sc i\kern-.025em b}\kern-.08em
    T\kern-.1667em\lower.7ex\hbox{E}\kern-.125emX}}
\begin{document}

\title{Evaluating Large Language Models Across 100+ Languages: A Unified Benchmark for Speech and Text Tasks}
\title{mSTEB: Massively Multilingual Evaluation of LLMs on Speech and Text Tasks\\
}

\author{\IEEEauthorblockN{Luel Hagos Beyene$^{1,2*}$, Vivek Verma$^{3,4*}$\thanks{*Equal contribution}, Min Ma$^5$, Jesujoba O. Alabi$^6$, Fabian David Schmidt$^7$,  \\ Joyce Nakatumba-Nabende$^8$, David Ifeoluwa Adelani$^{3,9,10}$} 
\IEEEauthorblockA{\textit{$^1$AIMS RIC, $^2$NM-AIST, $^3$Mila - Quebec AI Institute, $^4$Université de Montréal, $^5$Google DeepMind, $^6$Saarland University} \\ \textit{$^7$University of Würzburg, $^8$Makerere University, $^9$McGill University, $^{10}$Canada CIFAR AI Chair}.  \\ \\
Corresponding email: david.adelani@mila.quebec}
}

\maketitle

\IEEEpeerreviewmaketitle 
\begin{abstract}
Large Language models (LLMs) have demonstrated impressive performance on a wide range of tasks, including in multimodal settings such as speech. However, their evaluation is often limited to English and a few high-resource languages. For low-resource languages, there is no standardized evaluation benchmark. In this paper, we address this gap by introducing mSTEB, a new benchmark to evaluate the performance of LLMs on a wide range of tasks covering language identification, text classification, question answering, and translation tasks on both speech and text modalities. We evaluated the performance of leading LLMs such as Gemini 2.0 Flash and GPT-4o and state-of-the-art open models such as Qwen 2 Audio and Gemma 3 27B. Our evaluation shows a wide gap in performance between high-resource and low-resource languages, especially for languages spoken in Africa and Americas/Oceania. Our findings show that more investment is needed to address their under-representation in LLMs coverage. 
\end{abstract}

\begin{IEEEkeywords}
LLMs, AudioLLMs, machine translation, Speech translation, ASR, Topic Classification, QA, and NLI. 
\end{IEEEkeywords}

\section{Introduction}

Recently, Large language models (LLMs) have demonstrated impressive performance on various tasks~\cite{Achiam2023GPT4TR,DeepSeekAI2025DeepSeekR1IR,Dubey2024TheL3}, including in multimodal settings such as vision and speech~\cite{chu2023qwen,team2024gemini,tangsalmonn,Liu2023VisualIT,Yue2024PangeaAF}. However, their evaluation is often limited to the English language and a few other high-resource languages, such as Chinese and French. 
Many frontier LLMs claim multilingual capabilities yet rarely disclose the languages in their pre-training data. While extensive evaluations exist for text-based tasks across many languages~~\cite{adelani-etal-2024-sib,robinson2023chatgpt,ahuja2024megaverse}, comparable assessments for speech-based tasks, especially across hundreds of languages, remain scarce.



While speech model evaluations have expanded to many tasks, they often exclude LLMs and focus on only a few languages~\cite{arora2024evaluation}. Broad language coverage is mostly seen in language identification (LID) and automatic speech recognition (ASR), especially for speech representation models~\cite{chen2024towards,shi2024ml}. In contrast, recent audio LLMs like SALMONN~\cite{tangsalmonn} and Qwen 2 Audio~\cite{chu2023qwen} handle diverse tasks but support few languages. Some of these tasks such as audio captioning, scene classification, audio QA, and speech-to-text translation—are often more challenging than LID and ASR, especially for low-resource languages. Realistic multilingual LLM evaluation thus requires a broad range of tasks and languages, including low-resource ones.

In this paper, we introduce \msteb{}---a new benchmark for evaluating LLMs on a diverse set of tasks and languages based on aggregation of existing benchmarks. To ensure that it is massively multilingual, we cover over 100 languages for the speech-based tasks, and more than 200 languages for the text-based tasks. In both modalities, we cover LID, topic classification, translation and reading comprehension based question answering (QA). Additionally, we cover ASR for speech, and natural language inference for text. Our criteria for including an existing dataset is as follows: (1) they must cover at least one language in each of the following regions: Africa, Americas/Oceania, Asia or Europe. (2) they must be human labelled or created i.e. we do not include evaluation datasets that were machine translated and/or post-edited. 

Leveraging \msteb{}, we evaluated five state-of-the art LLMs:~\gpt{},~\gpta{}, ~\gemini{} (for both modalities),~\qwen{}, and~\gemma{}. The last two models are open-weight models while the others are proprietary. Our evaluation shows a large performance gap between the open models and proprietary models for both texts and speech tasks, however, the gap is much wider for the speech-based tasks, where~\qwen{} is worse on all languages. In contrast, in the text-based evaluations, we found~\gemma{} to be very competitive to the proprietary models on European and East Asian languages (CJK). Furthermore, there is a wide gap for the regions with a lot of low-resource languages such as Africa and Americas/Oceania regions. Our analysis demonstrates large gap in performance for many low-resource languages compared to high-resource languages, which shows that more efforts are needed to address their under-representation in LLMs.  We release our code, data and leaderboard on GitHub and HuggingFace platform.~\footnote{The code and datasets are available on \url{https://github.com/McGill-NLP/mSTEB}. We also release a leaderboard : \url{https://huggingface.co/spaces/McGill-NLP/msteb_leaderboard}}




\section{Related Work}

The landscape of multilingual modeling has shifted from classical representation learning models for text~\cite{conneau-etal-2020-unsupervised,he2021deberta,xue-etal-2021-mt5} and speech~\cite{babu22_interspeech,chen-etal-2024-towards-robust} to the development of LLMs with generative and cross-modal capabilities. In this new paradigm, a plethora of multilingual LLMs have been developed to support either~\cite{Achiam2023GPT4TR,Dubey2024TheL3}, or both modalities~\cite{nguyen-etal-2025-spirit}. These models are typically trained on large-scale multilingual and multimodal datasets sourced from the web. In parallel, several multilingual benchmark datasets have been created for both text~\cite{adelani-etal-2024-sib,nllbteam2022languageleftbehindscaling,Bandarkar_2024} and speech~\cite{Conneau2022FLEURSFL,Schmidt2025FleursSLUAM} tasks to evaluate their abilities. Although there has been a growing body of work on the evaluation of multilingual models, these efforts are often limited to a single modality and focus primarily on a small number of tasks and languages~\cite{keller2025speechtaxi}. As models continue to scale in both linguistic and multimodal coverage, it becomes increasingly important to evaluate and understand their progress--comparing them to specialized baselines and analyzing performance across different-language families and geographic regions. 
The current evaluation landscape thus reveals a significant opportunity for a benchmark that (1) comprehensively covers both speech and text modalities and make comparable analysis on the similar tasks using different modalities of input, (2) encompasses a diverse range of tasks relevant to modern LLM capabilities for both modalities, and (3) provides extensive language coverage, particularly for low-resource languages. \msteb{} is designed to address these gaps by providing a unified platform to evaluate LLMs across over 100 languages for speech tasks and over 200 for text tasks, including LID, topic classification, QA, and translation, thereby offering a more holistic view of their multilingual and multimodal prowess.

\section{mSTEB Benchmark}

We introduce \msteb---a \textbf{m}ultilingual \textbf{S}peech and \textbf{T}ext \textbf{E}valuation \textbf{B}enchmark covering five speech-based tasks and five text-based tasks. For the \textit{Speech-based} tasks, we cover LID, topic classification, reading comprehension QA (RC-QA), speech-to-text translation (S2TT) and ASR. While for the \textit{Text-based} tasks, we cover LID, topic classification, RC-QA, machine translation (MT) and natural language inference (NLI). All benchmarks are based on either Flores-200~\cite{nllbteam2022languageleftbehindscaling} or Fleurs~\cite{Conneau2022FLEURSFL} that are massively multilingual with additional annotations, except for NLI that we manually curated from HuggingFace datasets. We briefly describe the datasets below:


\textbf{Global NLI} is a \textit{new text-based benchmark} we created based on the aggregation of existing NLI datasets that are publicly available.  We combined several datasets consisting of 59 languages with a strong requirement that they are 
human-curated data either human translated from existing corpora such as the English MNLI~\cite{williams2018broad} or created manually for the language. The datasets we combined are the following: XNLI \cite{conneau-etal-2018-xnli}, AfriXNLI \cite{adelani2025irokobenchnewbenchmarkafrican} (African languages), IndicXNLI \cite{aggarwal-etal-2022-indicxnli} (Indian languages), AmericasNLI \cite{ebrahimi-etal-2022-americasnli}, XNLI-ca \cite{gonzalez-agirre-etal-2024-building-data} (Catalan), myXNLI \cite{lovenia-etal-2024-seacrowd} (Malaysian), IndoNLI \cite{mahendra-etal-2021-indonli} (Indonesia), JNLI \cite{kurihara-etal-2022-jglue} (Japanese), InferBR \cite{bencke-etal-2024-inferbr}, sick\_pl \cite{dadas-etal-2020-evaluation} (Polish), JamPatoisNLI \cite{DBLP:journals/corr/abs-2212-03419} (Jamaican Patois), KLUE \cite{park2021klue} (Korean), RoNLI \cite{Poesina-ACL-2024} (Romanian). For more fair comparison across languages, \textsc{Global NLI} dataset consists of 
a test set of 600 samples for each language. For all languages except Romanian, they have an equal number of entailment, neutral, and contradiction labels in validation and test sets. We also strive for equal representation of sentences from the 10 genres in the XNLI dataset. This was possible for 33 languages out of 59 coming from XNLI, AfriXNLI, IndicXNLI, XNLI-ca, and myXNLI datasets where we have the same premise and hypothesis across different languages.

\textbf{Flores-200} is a \textit{text-based} benchmark that was introduced by \cite{nllbteam2022languageleftbehindscaling} and is an extension of the original Flores-101\cite{goyal2022flores} dataset that covers 204 languages. It consists of sentences obtained from English Wikipedia articles, that have been professionally translated into 203 languages. We evaluate on the public \textit{devtest} set that consists of 1012 samples per language. We use this for the \textbf{machine translation (MT, en-xx and xx-en)} and the \textbf{LID} tasks. 

\textbf{SIB-200} is a \textit{text-based} benchmark introduced by \cite{adelani-etal-2024-sib} for \textbf{topic classification (TC)} task. It was created by manually annotating the English portion of the Flores-200~\cite{nllbteam2022languageleftbehindscaling} dataset, and then extending the annotation to the remaining 203 languages. Each sentence is classified into one of seven topics of classification: \textit{science/technology}, \textit{travel}, \textit{politics}, \textit{sports}, \textit{health}, \textit{entertainment}, and \textit{geography}. We evaluate on the \textit{test} set that consists of 204 samples per language.  

\textbf{Belebele} is a \textit{text-based} benchmark introduced by \cite{Bandarkar_2024} for multi-choice \textbf{(RC-QA)} for 122 languages. The data was created by asking annotators to create questions based on sentences in Flores, arranged back in document/paragraph level. Each of the questions is also accompanied by four multiple-choice answers. We evaluate on the public \textit{test} set that consists of 900 samples per language. 


\textbf{Fleurs} is a \textit{speech-based} benchmark 
that was introduced by \cite{Conneau2022FLEURSFL}, and it is a massively multilingual dataset covering 102 languages. It was extended from Flores-101~\cite{goyal2022flores} to include audio recordings of the human translated sentences. Each sentence was recorded by up to three participants. We leverage this dataset for the \textbf{LID}, \textbf{ASR} and  \textbf{S2TT} tasks (xx-en). 

\textbf{Fleurs-SLU} is a \textit{speech-based} benchmark that was recently introduced by \cite{Schmidt2025FleursSLUAM} by automatically aligning Fleurs audio utterances with SIB-200 for \textbf{TC} and Belebele for \textit{multi-choice} \textbf{RC-QA} tasks. The resulting datasets are known as SIB-Fleurs and Belebele-Fleurs respectively. 

Table \ref{tab:combined_dataset_statistics} provides a summary of all the tasks, datasets and the number of samples per language included in \msteb.

\begin{table}[t]
\centering
\caption{\textbf{Evaluation Data Statistics for mSTEB}}
\label{tab:combined_dataset_statistics}
\vspace{-1mm}
\setlength\tabcolsep{4pt}
\scalebox{0.90}{
\begin{tabular}{llccc}
\toprule
\textbf{Task}  & \textbf{Dataset} & \textbf{\#(Total Utts)} 
& \textbf{\#(Lang)} & \textbf{Per. Lang. Size} \\
\midrule

\multicolumn{5}{l}{\textbf{Speech Tasks}} \\
\midrule
LID & Fleurs & 77810 & 102 & 41--1041 \\ 
TC & SIB-Fleurs & 39270 & 102 & 18--527 \\ 
RC & Belebele & 65232 & 92 & 81--900 \\ 
S2TT (xx$\rightarrow$en) & Fleurs & 77163 & 101 & 41--1041 \\ 
S2TT (en$\rightarrow$xx) & Fleurs & 65347 & 101 & 647 \\ 
ASR & Fleurs & 77810 & 101 & 41--1041 \\ 
\midrule
\multicolumn{5}{l}{\textbf{Text Tasks}} \\
\midrule
LID & Flores-200 & 206448 & 204 & 1012 \\ 
TC & SIB-200 & 41820 & 205 & 204 \\ 
RC & Belebele & 109800 & 122 & 900 \\ 
MT & Flores-200 & 206448 & 204 & 1012 \\ 
NLI & Global NLI & 35400 & 59 & 600 \\ 

\bottomrule
\end{tabular}
}
\vspace{0.5em}
\end{table}

\section{Experimental setup}

\begin{table*}[ht]
\centering
\caption{\textbf{Prompts used for LLM evaluation}. For tasks in both modalities, we simply replace \textit{audio} in prompt by \textit{text}}
\label{tab:prompts_for_tasks}
\vspace{-1mm}
\setlength\tabcolsep{6pt}
\resizebox{\linewidth}{!}{%
\begin{tabular}{lp{18cm}}
\toprule
\textbf{Task} & \textbf{Prompt} \\
\midrule
ASR  & You are an assistant specialized in transcribing speech. Please generate an accurate transcript of the \{language\} audio provided below. Return only the transcription, no other text.  \\
\midrule
LID  & You are an assistant specialized in identifying spoken languages. Please identify the language in the audio clip below. Return only the name of the language, no other text.\footnotemark \\
\midrule
S2TT & \textbf{xx-en}: You are a translation expert. Listen to the following audio in \{source\_language\} and translate it to English. Return only the translated sentence. \\
\midrule
TC & You are an assistant able to classify topics in audios. \textbackslash n\textbackslash nGiven the categories Science/Technology, Travel, Politics, Sports, Health, Entertainment, or Geography; what is the topic of the \{language\} statement below? Return only the category, no other text.  \\
\midrule



RC & P: \{flores\_passage\}\textbackslash nQ: \{question\}\textbackslash nA: \{mc\_answer1\}\textbackslash nB: \{mc\_answer2\}\textbackslash nC: \{mc\_answer3\}\textbackslash nD: \{mc\_answer4\}\textbackslash nPlease choose the correct answer from the options above:\\
\midrule
MT & \textbf{xx-en}: You are a translation expert. Translate the following \{source\} sentences to English \textbackslash n\{source\} sentence: \{source\_column\}\textbackslash n\{target\} sentence: . Return only the translated sentence.\\
\midrule

NLI & Given the following premise and hypothesis in \{language\}, identify if the premise entails, contradicts, or is neutral towards the hypothesis. Please respond with exact 'entailment', 'contradiction', or 'neutral'. \textbackslash n\textbackslash nPremise: \{premise\} \textbackslash nHypothesis: \{hypothesis\}\\
\bottomrule
\end{tabular}
}
\end{table*}

\textbf{Baseline models: } For each of the task, we evaluate the performance of an existing supervised model, if available.
When there is no training data, 
we make use of cross-lingual transfer results as the baseline by fine-tuning on English training data, and evaluating on the remaining languages. 
For the \textit{speech-based tasks}, we make use of MMS-LID-2048 (trained on 2048 languages)~\cite{pratap2024scaling} for \textsc{LID}, 
\seamless{}~\cite{Barrault2025} for the ASR and S2TT tasks, and for the \textsc{TC} and \textsc{RC-QA}, we make use of the baseline from Fleurs-SLU paper~\cite{Schmidt2025FleursSLUAM} which is based on a cascaded approach---where audio utterances are first transcribed by an ASR/S2TT model (Seamless), followed by an evaluation on a text classifier (similar to \textit{Translate-Test}).
Similarly, for the \textit{text-based} tasks, we evaluate \textsc{GlotLID}~\cite{kargaran-etal-2023-glotlid} trained on over 2,000 languages on LID, NLLB-3.3B~\cite{nllbteam2022languageleftbehindscaling} on machine translation, 
Translate-Train
based on XLM-V~\cite{liang-etal-2023-xlm} for \textsc{RC-QA},  
and 
cross-lingual transfer of English model to other non-English languages for  \textsc{TC} and \textsc{NLI}. For the \tc{}, we make use of the NLLB-LLM2Vec encoder~\cite{schmidt2024self}---a model that stacks the encoder part of NLLB with the LLaMa-3 8B~\cite{llama3modelcard} decoder model to create an impressive multilingual sentence transformer model, based on LLM2Vec~\cite{behnamghader2024llmvec}, while for \nli{}, we leveraged mDeBERTaV3~\cite{he2023debertav}. For both \textsc{TC}, \textsc{RC-QA}, we make use of the English training data provided by the authors of the dataset, while for NLI, we fine-tuned the English MNLI~\cite{williams-etal-2018-broad} dataset. 

\textbf{LLMs evaluated: } We perform zero-shot evaluation on two popular open-weight models: \qwen{}~\cite{chu2023qwen} (speech) and \gemma{}~\cite{team2025gemma} (text). Additionally, we evaluate proprietary models:  \gpt{} (text), \gemini{} (text \& speech), and \gpta{} (speech). For all text-based tasks, except LID, we get the best prompt to use for evaluation from AfroBench \cite{ojo2025afrobenchgoodlargelanguage}. AfroBench tested three to five prompts for \textsc{TC} \textsc{RC-QA}, \textsc{MT}, and \textsc{NLI} on up to 55 languages. Given that we are working with over 200 languages and considering times and costs of operation, we chose the single best performing prompt for each task and run each task only once based on AfroBench recommendation. For LID, we kept a similar prompt across speech and text, changing a few words to identify the modality. For all languages, we provided the prompt in English, as this has been shown to work well for LLMs \cite{DBLP:journals/corr/abs-2112-10668}. The prompts used are provided in Table \ref{tab:prompts_for_tasks}. 


\textbf{Evaluation tools and hardware requirements:} 
We used the OpenAI API for GPT-4o (Audio) and Google Cloud API for Gemini-2.0 Flash. Gemma 3 27B ran on two NVIDIA L40S 48GB GPUs via vLLM~\cite{kwon2023efficient} for faster inference. For \qwen{}, we used two NVIDIA GeForce RTX 3090 for LID and topic classification, and a Quadro RTX 6000 \& Tesla V100S-PCIE-32GB for S2TT, RC-QA, and ASR.

\textbf{Evaluation metric: } We make use of accuracy metric for all classification tasks, Character-Error-Rate (CER) for the ASR, and ChrF++~\cite{popovic-2017-chrf} for MT and S2TT. We chose to use character-level metrics as they have been shown to be more reliable when evaluating diverse set of languages~\cite{kudugunta2023madlad} and better correlates with human judgment~\cite{freitag-etal-2024-llms}. We provide other metrics such as BLEU and WER in the supplementary material.

\textbf{Group by regions:} We categorize different languages into eight specific geographical regions reflecting the resource level between regions based on SIB-200~\cite{adelani-etal-2024-sib}. The languages are grouped into: Africa, Americas, Europe (W, S, N), Europe (E), Asia (W, C), Asia (S), Asia (SE), Asia (E), and Oceania; where W--``West", C-``Central", S--``South", E--``East", and N--``North".  We combined Americas and Oceania to form Americas/Oceania since the languages there are mostly the indigenous or Aboriginal languages.  

\section{Results And Discussions}

\begin{table*}[htbp]
\centering
\caption{\textbf{Overall Speech-based Evaluation results by region}. We used Qwen 2 7B (Qwen), Gemini-2-Flash (Gemini) and GPT-4o-Audio (12/24). Average (micro) for classification tasks left blank as they have different languages to classify.}
\label{tab:overall_by_region_speech}
\setlength\tabcolsep{4pt}
\resizebox{\linewidth}{!}{%
\begin{tabular}{lrrr|rrr|rrr|rrr|rrr|rrr}
\toprule
& \multicolumn{3}{c|}{\textsc{LID (acc $\uparrow$)}} &  \multicolumn{3}{c|}{\textsc{Topic class. (acc $\uparrow$)}}  &  \multicolumn{3}{c|}{\textsc{RC-QA (acc $\uparrow$)}}  &   \multicolumn{3}{c|}{\textsc{ASR (CER $\downarrow$)}} &   \multicolumn{3}{c|}{\textsc{S2TT (ChrF++ $\uparrow$)}} & \multicolumn{3}{c}{\textsc{Ave. (class. tasks)}}\\

\textbf{Region} & \rotatebox{40}{\textbf{Qwen}} &  \rotatebox{40}{\textbf{GPT-4o}} & \rotatebox{40}{\textbf{Gemini}} & \rotatebox{40}{\textbf{Qwen}} &  \rotatebox{40}{\textbf{GPT-4o}} & \rotatebox{40}{\textbf{Gemini}} & \rotatebox{40}{\textbf{Qwen}} &  \rotatebox{40}{\textbf{GPT-4o}}  & \rotatebox{40}{\textbf{Gemini}} & \rotatebox{40}{\textbf{Qwen}} & \rotatebox{40}{\textbf{GPT-4o}} & \rotatebox{40}{\textbf{Gemini}} & \rotatebox{40}{\textbf{Qwen}} & \rotatebox{40}{\textbf{GPT-4o}}  & \rotatebox{40}{\textbf{Gemini}} &  \rotatebox{40}{\textbf{Qwen}} & \rotatebox{40}{\textbf{GPT-4o}}  & \rotatebox{40}{\textbf{Gemini}}  \\
\midrule				
Africa      & 0.0  & 56.6 & \bf 68.5 
            & 28.2 & 55.3 & \bf 68.5 
            & 26.3 & 45.6 & \bf 55.5 
            & 83.9 & 60.1 & \bf 18.0  
            & 13.2 & 28.5 & \bf 37.6
            & 18.2 & 52.5 & \bf 64.2 \\
Americas/Oceania    & 0.3  & \bf 98.1 & 96.7 
                    & 15.2 & \bf 66.9 & 44.7 
                    & 27.1 & 45.6 & \bf 50.9 
                    & 146.2 & 31.7 & \bf 19.5 
                    & 11.9 & 36.5 & \bf 27.2
                    & 14.2 & 70.2 & \bf 64.1 \\
Asia (S)  & 4.5  & 91.8 & \bf 97.5 
            & 32.8 & 79.8 & \bf 83.0 
            & 30.3 & 62.2 & \bf 71.5 
            & 113.7 & 35.1 & \bf 7.7  
            & 16.0 & 47.3 & \bf 58.3
            & 22.5 & 78.0 & \bf 84.0 \\
Asia (SE)  & 10.5 & 88.8 & \bf 91.0 
            & 33.0 & 76.0 & \bf 82.3 
            & 31.4 & 63.8 & \bf 75.8 
            &  97.8 & 35.0 & \bf 7.3   
            & 16.0 & 48.0 & \bf 57.8
            & 24.9 & 76.2 & \bf 83.0 \\
Asia (W, C)  & 21.7 & 80.2 & \bf 88.0 
             & 28.4 & 79.1 & \bf 82.1 
            & 30.1 & 64.6 & \bf 74.7 
            & 111.7 & 27.2 & \bf 6.7    
            & 15.6 & 47.5 & \bf 55.5
            & 26.7 & 74.6 & \bf 81.6 \\
Asia (E)  & 59.2 & 90.5 & \bf 98.1 
            & 58.7 & 80.5 & \bf 83.5 
            & 43.5 & 70.0 &\bf 77.3 
            & 48.2 & 23.2 & \bf 13.9 
            & 34.0 & 47.2 & \bf 52.8
            & 53.8 & 80.3 & \bf 86.3 \\
Europe (W, N, S) & 28.6 & \bf 86.1 & 84.5 
            & 54.5 & \bf 84.0 & 82.1 
            & 41.6 & 74.3 & \bf 81.7 
            & 62.6 & 12.4 & \bf 7.3  
            & 29.6 & 56.9 & \bf 58.3
            & 41.5 & 81.5 & \bf 82.8 \\
Europe (E) & 11.8 & 90.4 & \bf 94.1 
            & 53.6 & \bf 85.6 & 84.9 
            & 36.6 & 74.0 &\bf 82.1 
            & 98.2 & 12.0 & \bf 9.9
            & 23.9 & 59.3 & \bf 62.2
            & 34.0 & 83.3 & \bf 87.0 \\
\midrule   
Average (macro)  & 17.1 & 85.3 & \bf 89.8 
                & 38.0 & 75.9 & \bf 76.4 
                & 33.4 & 62.5 & \bf 71.2 
                & 95.3 & 29.6 &  \bf 11.3 
                & 20.0 & 46.4 & \bf 51.2
                & 29.5 & 74.6 & \bf 79.1 \\
Average (micro)  & 15.8 & 81.2 & \bf 86.1
                & 40.4 & 76.0 & \bf 79.5 
                & 33.3 & 63.4 & \bf 72.6 
                & 88.4 & 30.1 & \bf 10.3 
                & 20.5 & 47.3 & \bf 53.5
                &      &     \\

\bottomrule
\end{tabular}
}
\end{table*}

\begin{table*}[t]
\centering
\caption{\textbf{Overall Text-based Evaluation results by region}. We used Gemma 3 27B (Gemma), Gemini-2-Flash (Gemini) and GPT-4o (08/24). Average (micro) for classification tasks left blank as they have different languages to classify.}
\label{tab:overall_by_region_text}
\setlength\tabcolsep{4pt}
\resizebox{\linewidth}{!}{%
\begin{tabular}{lrrr|rrr|rrr|rrr|rrr|rrr}
\toprule
 & \multicolumn{3}{c|}{\textsc{LID (acc $\uparrow$)}} &  \multicolumn{3}{c|}{\textsc{Topic class. (acc $\uparrow$)}}  &  \multicolumn{3}{c|}{\textsc{RC-QA (acc $\uparrow$)}}  &   \multicolumn{3}{c|}{\textsc{NLI (acc $\uparrow$)}} &   \multicolumn{3}{c|}{\textsc{Machine Transl. (ChrF++ $\uparrow$)}} & \multicolumn{3}{c}{\textsc{Ave. (class. tasks)}}\\
\textbf{Region} & \rotatebox{60}{\textbf{Gemma}} &  \rotatebox{60}{\textbf{GPT-4o}} & \rotatebox{60}{\textbf{Gemini}} & \rotatebox{60}{\textbf{Gemma}} &  \rotatebox{60}{\textbf{GPT-4o}} & \rotatebox{60}{\textbf{Gemini}} & \rotatebox{60}{\textbf{Gemma}} &  \rotatebox{60}{\textbf{GPT-4o}} & \rotatebox{60}{\textbf{Gemini}} & \rotatebox{60}{\textbf{Gemma}} &  \rotatebox{60}{\textbf{GPT-4o}} & \rotatebox{60}{\textbf{Gemini}} & \rotatebox{60}{\textbf{Gemma}} &  \rotatebox{60}{\textbf{GPT-4o}} & \rotatebox{60}{\textbf{Gemini}} & \rotatebox{60}{\textbf{Gemma}} &  \rotatebox{60}{\textbf{GPT-4o}} & \rotatebox{60}{\textbf{Gemini}} \\
\midrule				
Africa              & 62.9 & 69.6           & \textbf{80.0} & 59.9 & 69.6          & \textbf{74.9} & 60.6 & 70.2          & \textbf{77.0} & 58.4          & 61.8          & \textbf{64.2} & 34.8/24.0 & 41.0/30.5 & \textbf{46.2/32.0} & 60.4 &          67.8 & \textbf{74.0} \\ 

Americas/Oceania    & 79.2 & \textbf{99.6}  & 99.5          & 69.5 & 80.5          & \textbf{82.9} & 62.3 & 78.2          & \textbf{80.7} & 46.1          & \textbf{49.1} & 46.8          & 41.2/32.4 & 49.0/40.8 & \textbf{52.6/42.9} & 64.3 &          76.9 & \textbf{77.5} \\ 

Asia (S)            & 78.2 & 83.2           & \textbf{87.2} & 81.5 & 82.1          & \textbf{86.7} & 78.6 & 83.0            & \textbf{84.7} & 71.1          & \textbf{72.4} & 71.4          & 51.9/33.3 & 54.2/34.2 & \textbf{59.6/39.2} & 77.4 &          80.2 & \textbf{82.5} \\ 

Asia (SE)           & 76.2 & 82.0           & \textbf{86.6} & 75.8 & 80.9          & \textbf{84.5} & 75.5 & 76.3          & \textbf{81.7} & 76.2          & \textbf{77.7} & 75.4          & 48.4/35.1 & 51.8/38.5 & \textbf{57.6/42.1} & 75.9 &          79.2 & \textbf{82.1} \\ 

Asia (W, C)         & 99.3 & 99.2           & \textbf{99.5} & 85.3 & 86.8          & \textbf{87.7} & 83.3 & 87.0           & \textbf{87.5} & 75.0           & \textbf{77.2} & 75.2          & 56.5/38.8 & 58.6/44.3 & \textbf{60.8/45.9} & 85.7 & \textbf{87.6} &         87.5 \\ 

Asia (E)            & 99.7 & \textbf{100.0}   & 99.8          & 85.0 & \textbf{87.2} & 86.7          & 87.7 & \textbf{90.2} & 85.9          & 75.2          & \textbf{81.2} & 76.6          & 54.5/29.7 & 56.3/\textbf{33.7} & \textbf{57.4}/32.8 & 86.9 & \textbf{89.6} &         87.2 \\ 

Europe (W, N, S)    & 94.2 & \textbf{98.4}  & 97.9          & 85.1 & 87.5          & \textbf{88.2} & 90.3 & \textbf{92.6} & 87.2          & 82.7          & \textbf{84.8} & 83.6          & 62.9/52.7 & 64.9/56.5 & \textbf{66.2/56.9 }& 88.1 & \textbf{90.8} &         89.2 \\ 

Europe (E)          & 90.0 & 95.7           & \textbf{97.8} & 85.0 & 87.5          & \textbf{88.0}   & 89.2 & \textbf{92.8} & 86.2          & \textbf{77.0}   & 75.8 & 75.7   & 60.3/48.5 & 62.4/49.2 & \textbf{63.9/51.1} & 85.3 & \textbf{88.0} & 86.9 \\ 

\midrule       
Average (macro)     & 85.0 & 91.0           & \textbf{93.5}  & 78.4 & 82.8          & \textbf{85.0}   & 78.4 & 83.8   & \textbf{83.9} & 70.2          & \textbf{72.5} & 71.1 & 51.3/36.8 & 54.8/41 & \textbf{58/42.9} & 78.0 & 82.5 & \textbf{83.4} \\ 
Average (micro)     & 81.0 & 86.4           &\textbf{90.5}  & 75.8 & 80.6          & \textbf{83.5} & 77.5  &82.3   & \textbf{83.4} & 65.4          & \textbf{67.9} & 67.3 &49.7/36.4   &53.3/40.7 &\textbf{57/42.7} &   \\

\bottomrule
\end{tabular}
}

\end{table*}

\subsection{Overall Results}

\begin{figure}[t]
  \centering
  \begin{subfigure}[b]{0.23\textwidth}
    \centering
    \includegraphics[width=\textwidth]{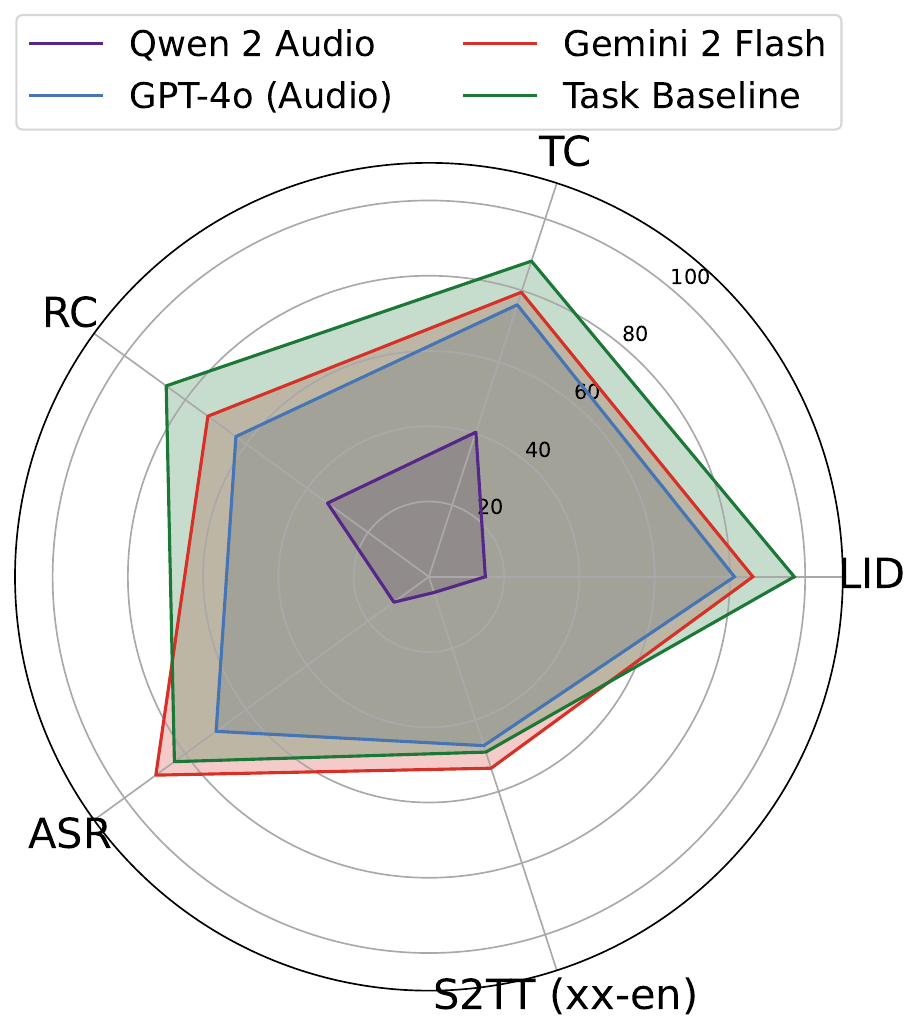}
    \caption{Speech}
    \label{fig:first}
  \end{subfigure}
 \hspace{-0.75\baselineskip}
 \begin{subfigure}[b]{0.25\textwidth}
    \centering
    \includegraphics[width=\textwidth]{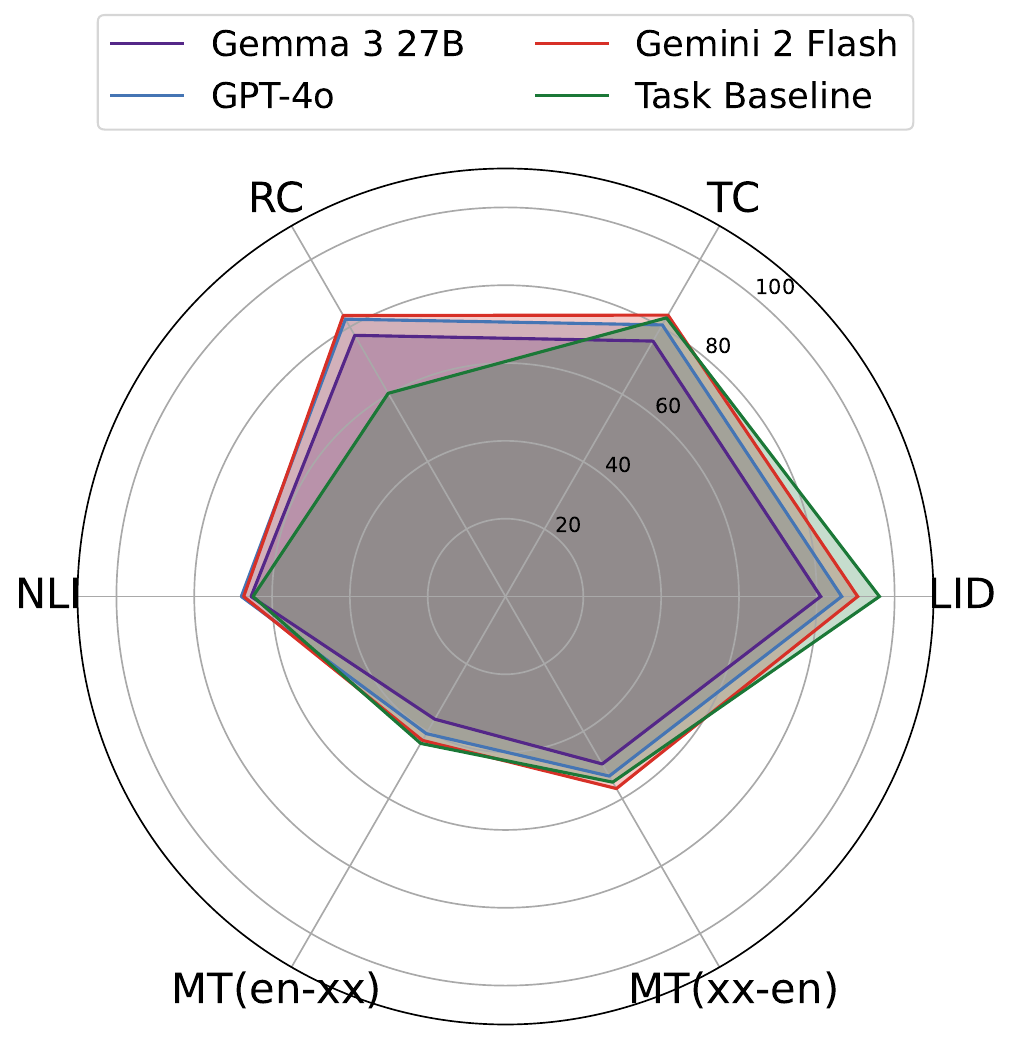}
    \caption{Text}
    \label{fig:second}
  \end{subfigure}
  \caption{\textbf{Average results by tasks}. For the ASR, we reported ``$100-CER$'' value. }
  \label{fig:performance_tasks}
\end{figure}

\subsubsection{Results by Task} Figure \ref{fig:performance_tasks} shows the result of various LLMs and the best result obtained from a task-specific model (i.e. \textit{Task Baseline}). For the \textbf{speech tasks}, we find that for classification tasks (i.e. \lid{}, \tc{}, and \rc{}), our result shows that the task baseline achieved much better result, for example, the MMS-LID accuracy is more than $+10$ points better than the best LLM (\gemini{}). Similarly, for \tc{} and \rc{}, the performance gap is more than $+8$ and $+13$ points respectively. Overall, we saw a wide performance gap between \qwen{} and \gemini{} which clearly shows that the former only covers a few languages. For more challenging tasks such as \asr{} and \st{}, surprisingly, we found \gemini{} to be better than \seamless{}, one of the SOTA task-specific baseline for both tasks. This result shows that \gemini{} should be preferred for these speech tasks especially for low-resource languages rather than their task-specific variants. For the \textbf{text-based tasks}, the finding is different, we find that the gap in performance between an open weight model such as \gemma{} and \gemini{} or \gpt{} is much smaller. Similarly, the performance of task baselines are also very similar to those of the LLMs on almost all tasks except for \lid{} and \rc{}. For \lid{}, task baselines (GlotLID) is better by $+5$ points accuracy when compared to \gemini, while for \rc{}, all LLMs are significantly better than task baseline based on Translate-TRAIN (on XLM-V) with over $+17$ points.  

\subsubsection{Results by Region}
\label{sec:result_by_region}
Table ~\ref{tab:overall_by_region_speech} and \ref{tab:overall_by_region_text} shows the result across eight different \textit{geographical groups}. For the \textbf{speech tasks}, we find that the \textit{African} and the \textit{Americas/Oceania} region achieved the worst result across all the covered regions probably because most of the low-resource languages are concentrated there. While \qwen{} model achieved the worst results among the LLMs we evaluated, we find that its performance is slightly better for Asia (E) and Europe (W, N, S) which may reflect where most of the models pre-training data are coming from. For example on \asr{} and \st{}, it achieved less than 50 CER and more than 30 ChrF on East Asia region. \gemini{} on the other hand had less than 15 CER for all regions except \textit{Americas/Oceania} and \textit{Africa}. However, \gpt{} was worse than \gemini{} across all regions we tested, it only achieved less than 15 CER for the European regions. The worst performance of \gpt{} is also for African languages at 60.1 CER. We observe similar trend for \st{} where \gemini{} achieved more than 50 ChrF++ on all regions except the \textit{Americas/Oceania} and \textit{African} regions. For \textit{classification tasks}, \gemini{} had over 70 accuracy points on \tc{} and \rc{} except for the African and Americas region; however, for the \lid{}, the most language confusion is for the African region.  
For \textbf{text-based tasks}, results were generally high except in Africa. While we observe the Americas region also has high performance across tasks including MT, this is not true for NLI where it achieved the lowest ($46.8$), more than $-17$ points behind Africa. A possible reason is that large datasets like Flores and SIB-200 cover few Americas/Oceania languages (about five, including two Creole languages), whereas AmericasNLI includes more truly low-resource ones (e.g., Otomi, Wixarika), offering a more realistic picture of performance of indigenous Americas languages.


\subsubsection{Results by Language Family} In Table ~\ref{tab:gemini_2_lang_family}, we show the performance by language family comparing tasks that are covered in both speech and text modalities. We cover four tasks: \lid{}, \tc{}, \rc{}, and translation (MT \& S2TT) in the \textit{xx-en} direction, and excluded languages not covered in both modalities. Thus, more than 90 languages from the text modality are missing. 

\textbf{Text-based tasks achieve overall higher performance: } In general, performance on texts is higher than performance on speech for same tasks. This implies that current LLMs are better on texts than speech, which implies that more investment in developing better audio LLMs are needed. On the LID task, several language families especially those that use a different script have perfect accuracy score using \gemini{}, however, we noticed some mis-classifications for the spoken utterances. 
Similar trend is observed for \tc{} and \rc{} classification tasks; on \tc{}, all language families except \textit{Nilotic} achieved more than $80$ accuracy point. Also, for the translation tasks, several language families achieved impressive ChrF++ of $60$ points and above in \textit{xx-en} direction while only one language family (Uralic) achieved up to 60 ChrF++ in S2TT.  

\textbf{Language families in Africa achieved the lowest performance: } For both text and speech modalities, the language families with the lowest performance are: \textit{Atlantic-Congo}, \textit{Nilotic}, and \textit{Afro-Asiatic}, many or all of the languages in the first two language families are often categorized as low-resourced according to Joshi's classification~\cite{joshi-etal-2020-state} and are only located in Africa. This confirms our findings in \textbf{Sec.~\ref{sec:result_by_region}}. Apart from having low performance, the gap in performance is also quite wide. For example, the average ChrF++ for \st{} in \textit{Nilotic} and \textit{Atlantic-Congo} is $-36.5$ and $-25.9$ lower when compared to Indo-European languages. Similarly, for \rc{}, the performance of Indo-European is $78.4$ while \textit{Atlantic-Congo}, \textit{Nilotic}, and \textit{Afro-Asiatic} achieved $-36.5$, $-25.9$, and $-25.9$ respectively. 

In addition to the language families in Africa, we find other language families for text modalities that often achieve lower performance. A few language families are in the Americas region such as Quechuan and Aymaran. Similarly, there are Austroasiatic languages in the South-East Asia such as Khmer and Santali. This findings highlight that the importance of scaling LLMs to more languages in the Africa, South-East Asia and Americas/Oceania regions.

\begin{table}[t]
\centering
\caption{\textbf{Gemini-2.0-Flash results across all Language Families and Modalities for similar Sets of Tasks}}
\label{tab:gemini_2_lang_family}
\setlength\tabcolsep{4pt}
\resizebox{\linewidth}{!}{%
\begin{tabular}{lrrrr|rrrr}
\toprule
& \multicolumn{4}{c|}{\textsc{Speech}} &  \multicolumn{4}{c}{\textsc{Text}}  \\

& \textbf{LID} & \textbf{TC} 
& \textbf{RC}  & \textbf{S2TT} & \textbf{LID} & \textbf{TC} & \textbf{RC} & \textbf{MT}   \\

\textbf{Lang. Family}  & \textit{acc} &\textit{acc} 
&\textit{acc} & \textit{ChrF++}  &\textit{acc}  & \textit{acc} & \textit{acc}  & \textit{ChrF++}   \\

\midrule				
      Indo-European & 86.7 & 82.9 & 78.4 & 59.1 & 99.4 & 88.3 & 87.8 & 65.2 \\ 
        Atlantic-Congo & 63.6 & 65.9 & 52.4 & 33.2 & 92.4 & 80.7 & 73.4 & 51 \\ 
        Afro-Asiatic & 95 & 76.4 & 66.9 & 49.4 & 100 & 86.1 & 84.6 & 62.7 \\ 
        Austronesian & 89.1 & 76.7 & 73 & 56.6 & 98.4 & 87.4 & 88.4 & 65.3 \\ 
        Turkic & 92 & 83.5 & 74.7 & 54.9 & 99.8 & 87.2 & 87.7 & 58.9 \\ 
        Sino-Tibetan & 96.8 & 81.9 & 75 & 48 & 99.8 & 86.6 & 86.7 & 56.6 \\ 
        Nilotic & 49.6 & 55.6 & 40 & 22.6 & 99.2 & 74 & 58 & 39.9 \\ 
        Dravidian & 99.6 & 81.9 & 70.5 & 58 & 100 & 87.3 & 84.6 & 62.1 \\ 
        Tai-Kadai & 86.1 & 82.5 & 74.4 & 56.6 & 100 & 87 & 83.8 & 61.2 \\ 
        Uralic & 99 & 84.3 & 80.9 & 61.3 & 100 & 88.1 & 73.5 & 63.1 \\ 
        Austroasiatic & 100 & 81.3 & 75.1 & 54.4 & 100 & 88.2 & 87.9 & 60.8 \\ 
        Japonic & 100 & 84.3 & 78.5 & 54.5 & 100 & 89.2 & 87.8 & 56 \\ 
        Koreanic & 100 & 84.5 & 83 & 57.4 & 100 & 85.3 & 82.7 & 58.3 \\ 
        Mongolic-Khitan & 100 & 83 & 72 & 51.9 & 100 & 84.8 & 85.1 & 58.5 \\ 
        Kartvelian & 99.9 & 83.5 & 76.9 & 55.2 & 100 & 88.2 & 88.6 & 57.2 \\ 
\midrule
Average ($\uparrow$) & 90.5 & 79.2 &71.4 &  51.5  & 98.5 & 86.7 & 84.6 & 61.6\\ 
\bottomrule
\end{tabular}

}

\end{table}

\subsection{Detailed analysis of task specific results }
\paragraph{LID}

\begin{figure*}[t]
  \centering
  \begin{subfigure}[b]{0.355\textwidth}
    \centering
    \includegraphics[width=\textwidth]{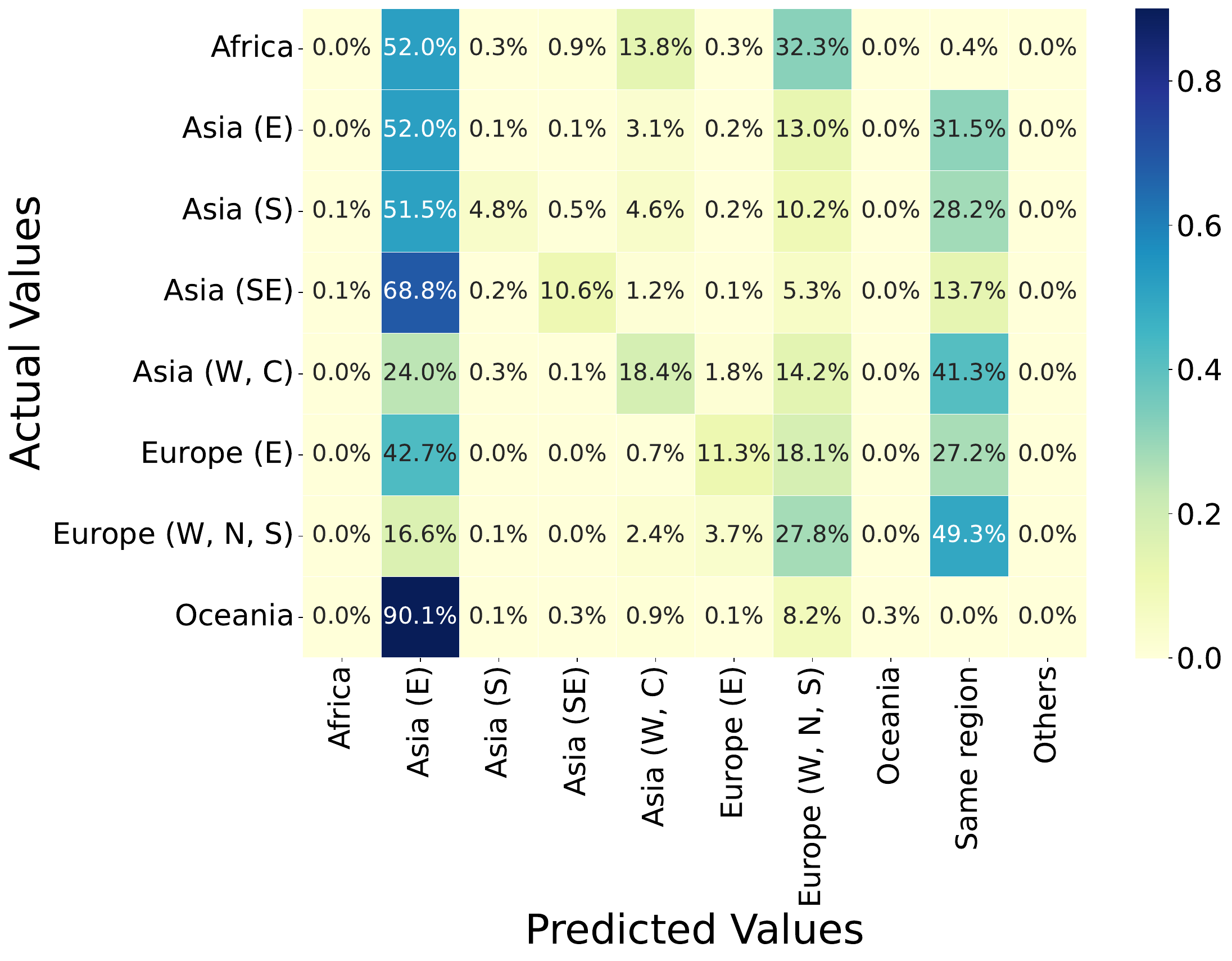}
    \caption{\qwen{}}
    \label{fig:first}
  \end{subfigure}
  \hspace{-1.95\baselineskip}
  \begin{subfigure}[b]{0.355\textwidth}
    \centering
    \includegraphics[width=\textwidth]{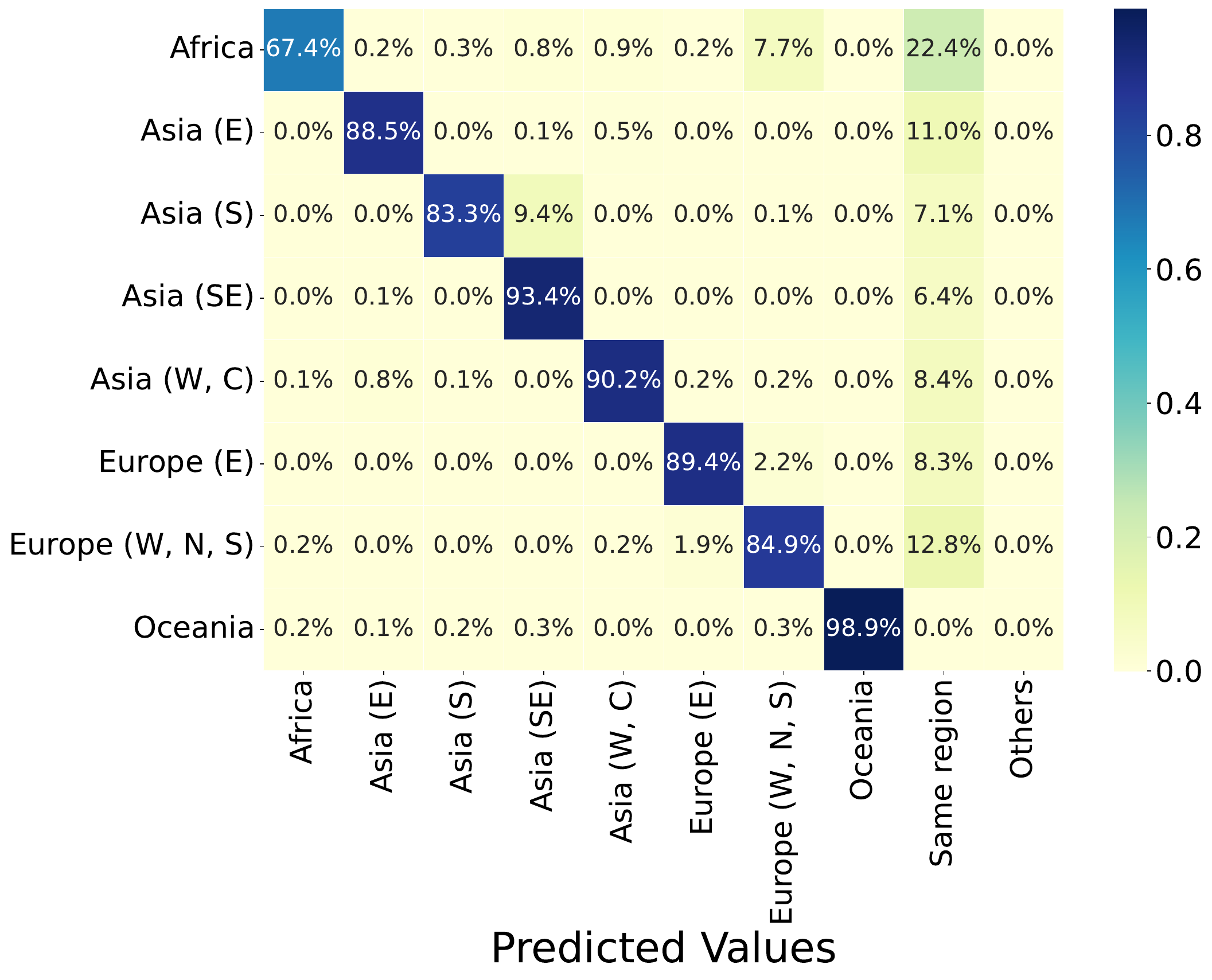}
    \caption{\gpta{}}
    \label{fig:second}
  \end{subfigure}
  \hspace{-1.95\baselineskip}
  \begin{subfigure}[b]{0.355\textwidth}
    \centering
    \includegraphics[width=\textwidth]{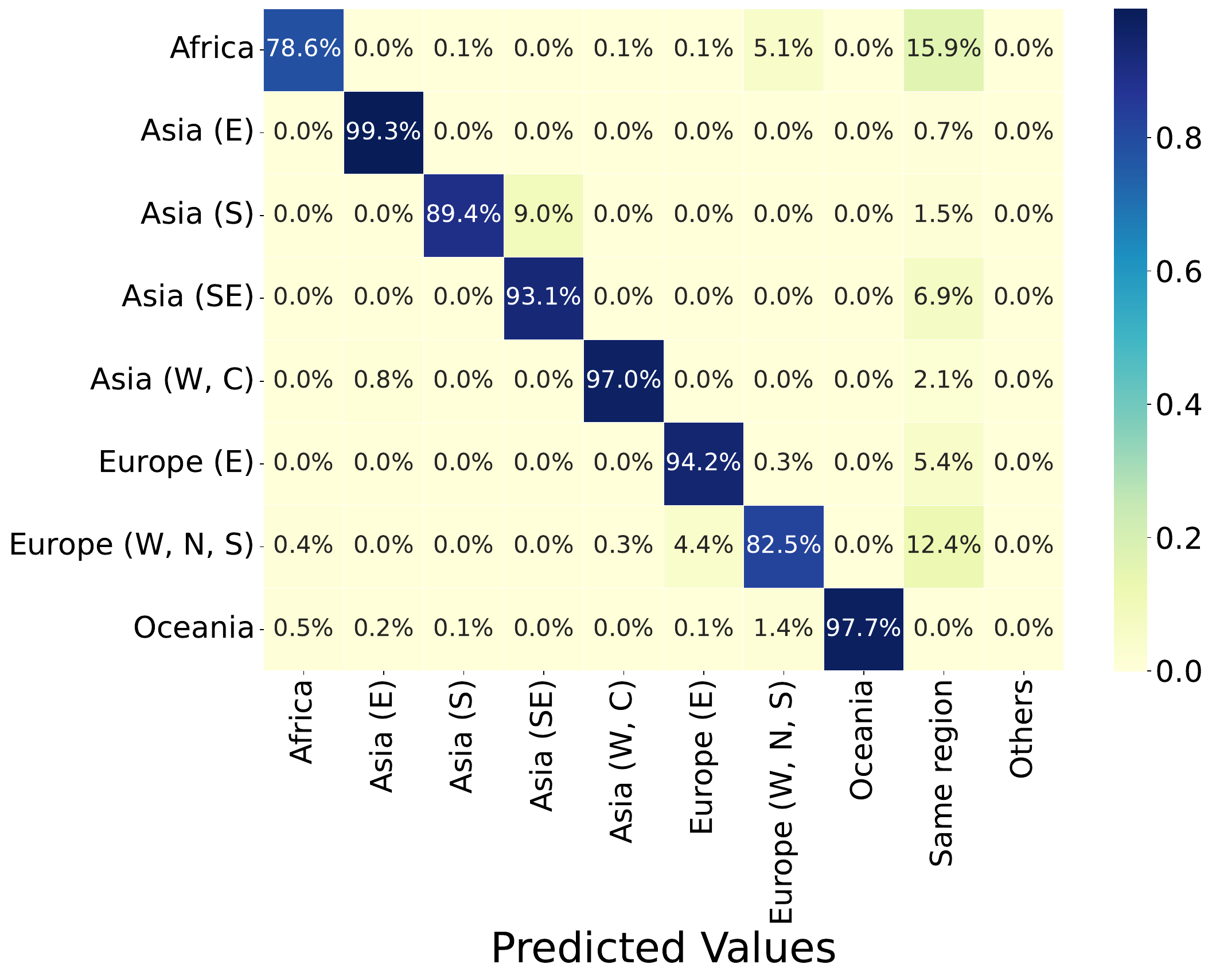}
    \caption{\gemini{}}
    \label{fig:third}
  \end{subfigure}
  \vspace{-3mm}
  \caption{Confusion matrix for LID task per system, where the predictions over 102 languages had been aggregated into corresponding regions for visualization.}
  \label{fig:lid_confusion}
\end{figure*}

As shown in the Table I and II, \gemini{} performed better than \gpt{} and \gemma{}, in both textual and speech language identification tasks, especially in African, Southern and Southeastern Asian languages. For text LID, the rankings of most challenging language groups to classify are: Africa \textgreater{} Southeastern Asia \textgreater{} Southern Asia \textgreater{} Americas/Oceania \textgreater{} Most European languages, while the Eastern, Western and Central Asian languages seems easy to be distinguished with nearly perfect accuracy.



In Figure ~\ref{fig:lid_confusion}, we plot the confusion matrix between for the three audio LLMs across regions. For languages with wrong prediction, but the predicted language is still in the same region, we labelled them as \texttt{Same region}, while the predicted languages not covered in the Fleurs dataset, we assigned \texttt{Others}. 
As observed, \qwen{} tended to mis-classify languages of various other regions into Asian languages, and confused many European languages with African languages. \gpta{} and \gemini{} models significantly perform better, even when misclassification occured, it usually happened within the same geographical regions. For example, in the \textit{Africa} region with the lowest performance, 22.4\% languages were predicted incorrectly but within the same region for \gpta{}, while \gemini{} did better with a 15\% error rate for same region mis-classification. Next to Africa is \textit{Europe (W, N, S)}, where the most language confusion is either in the same region or with Eastern European languages. On a closer look on the least performant Europe (W, N, S) languages,  \textit{Occitan} and \textit{Asturian} have the least performance with less than $20\%$ accuracy points on \gpta{} and \gemini{}, followed by luxembourgish achieving around 40\% on both proprietary models. Individual differences also exist, we find \gemini{} struggles more with \textit{Irish} at $66.5\%$ and \textit{Bosnian} at $12.3\%$ while \gpta{} struggles with \textit{Bosnian} at $68.9\%$ and \textit{Galician} at $76.1\%$.

\begin{figure}[t]
    \centering
    \includegraphics[width=0.47\textwidth]{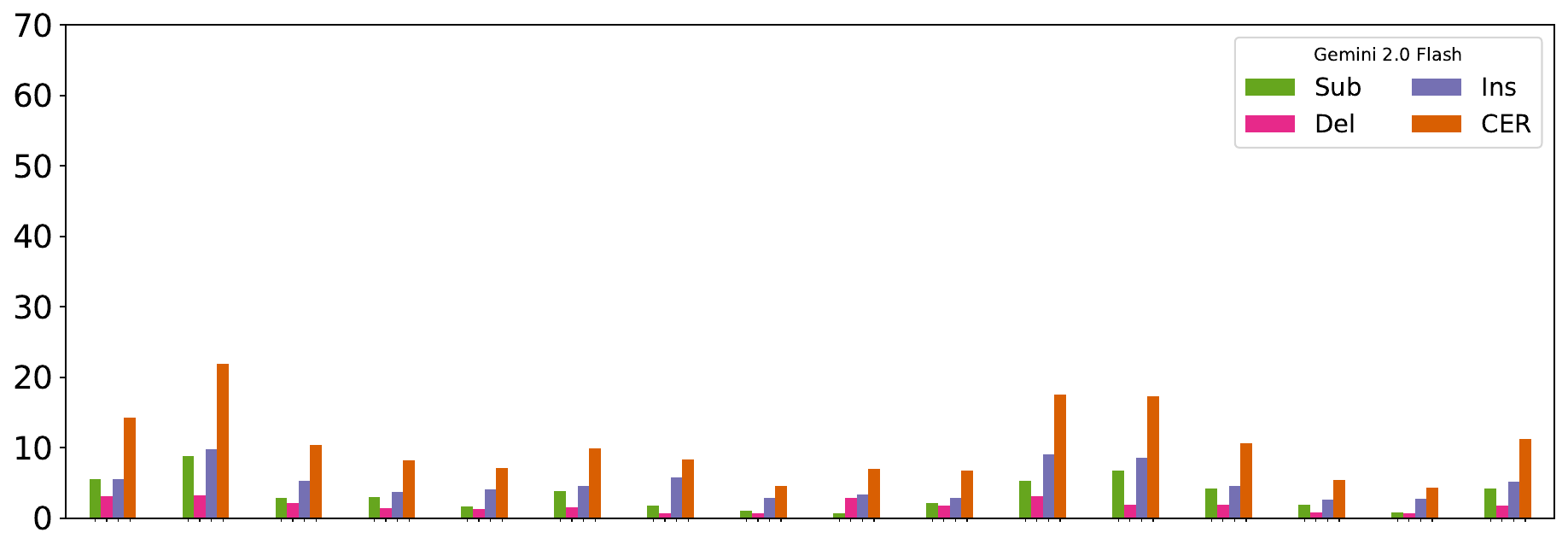}
    \includegraphics[width=0.47\textwidth]{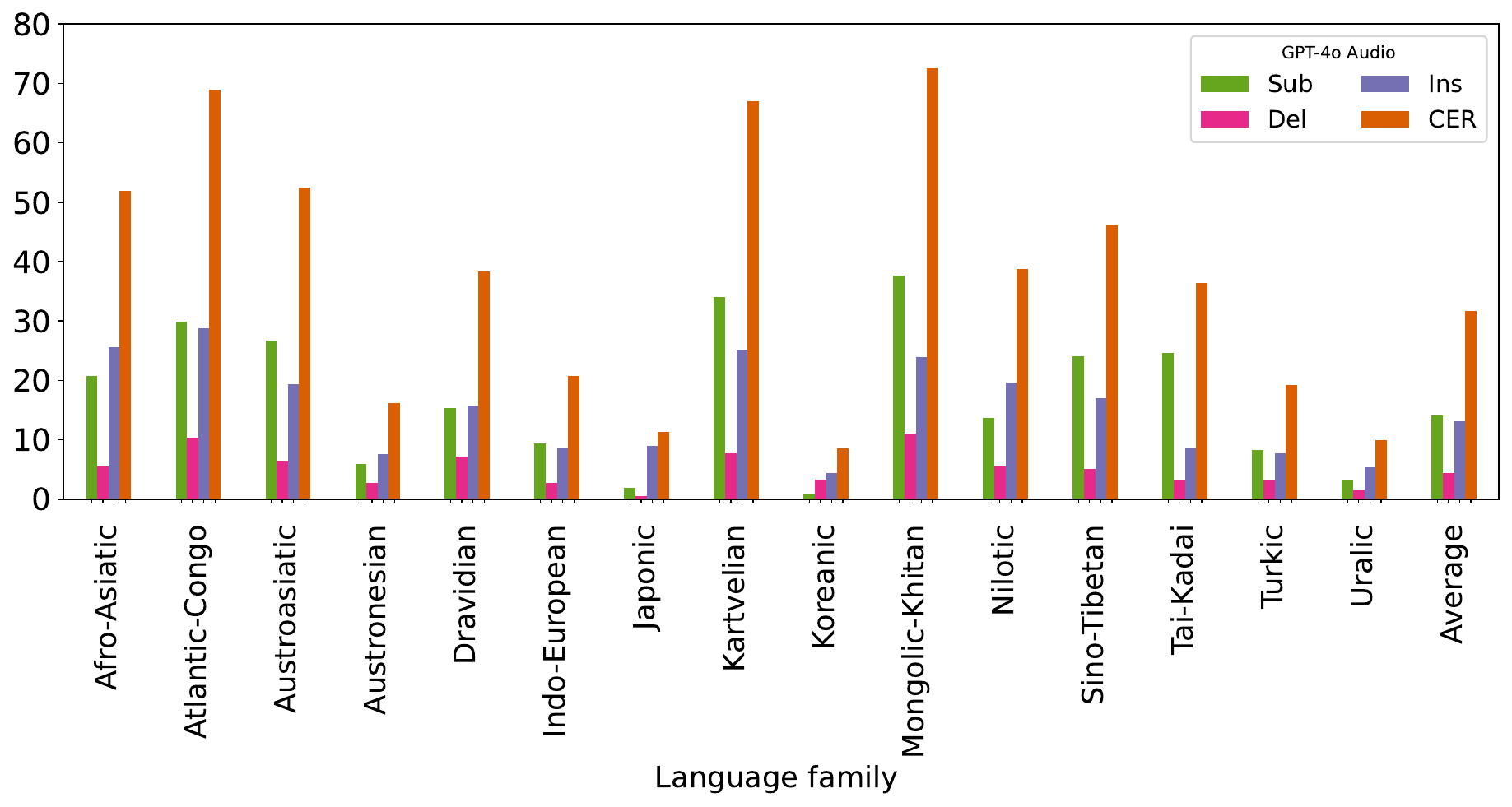}
     \caption{\textbf{Fine-grained error analysis for proprietary LLMs}}
     \label{fig:gpt4o_asr_error}
    \vspace{-5mm}
\end{figure}

\paragraph{ASR}
On the task of ASR, \gemini{} substantially outperformed \qwen{} and \gpta{}, except in Eastern Asian languages. 
We list the full WER numbers per language in the supplement material. For detailed error types, we show them in Figure~\ref{fig:gpt4o_asr_error}, aggregated by language families.  Substitution and insertion errors constitute the primary ASR error patterns for both \gpta{} and \gemini{}. The models diverge in their principal error types, with \gemini{} prone to more insertions and \gpta{} to more substitutions. The highest incidence of insertion errors for \gemini{} occurred in the Atlantic-Congo, Mongolic-Khitan, and Nilotic language families. For \gpta{}, substitution errors were most common in the Mongolic-Khitan, Kartvelian, and Atlantic-Congo families. In terms of strengths, \gemini{} performed best with Kartvelian, Uralic, and Turkic languages. \gpta{}, despite a high CER in Kartvelian, showed strong performance in Koreanic, Japonic, and Uralic. Both models, however, faced significant challenges with the Afro-Asiatic and Atlantic-Congo languages.

\paragraph{NLI}
The Global NLI dataset has the highest representation of Americas/Oceania amongst all the other datasets that we evaluated, with 11 out of the 59 languages being from this region. This is due to the Americas NLI dataset \cite{ebrahimi-etal-2022-americasnli} and only 3 languages of this region are present in the other text-based tasks. On this task, \gpt{} performs the best overall and in most regions. Americas/Oceania is the region with the lowest performance in this task. We see a performance gap of $-35$ points between Americas/Oceania (49.1\%) and Europe (W, N, S) (84.8\%), which is larger than that present in other tasks. 
The NLI results show that many Americas/Oceanian languages are less covered by many LLMs, although only few multilingual datasets includes these indigeneous languages. More efforts are needed to expand benchmarks for this region of languages. 



\section{Conclusion}
In this paper, we introduced \msteb{}---a new massively multilingual benchmark covering five tasks in each of text and speech modalities across more than 200 languages. We evaluate the performance of two open weight models and three proprietary models on these tasks. Our findings show a wide gap in performance between open models and proprietary models for the speech modality while in the text modality, the gap is very small. Similarly, we find multimodal LLMs to be better than their task specific baselines on more challenging speech tasks such as ASR and S2TT. Finally, we find that the languages with the lowest performance in both modalities are from the regions of Africa and Americas/Oceania. We hope our findings will encourage more investment in the development of multimodal LLMs for these regions. 

\section*{Acknowledgment}
This research was supported by the Google grant via Mila and the IVADO R3AI Regroupement
Grant. Luel acknowledges the support of Carnegie Corporation of New York provided through AIMS-RIC. We would also like to thank Google Cloud for the GCP credits Award through the Gemma 2 Academic Program, and OpenAI for providing us access for providing API credits. 

\bibliographystyle{IEEEtran}
\bibliography{IEEEfull}

\end{document}